\documentclass{sig-alternate-05-2015}
\usepackage{graphicx}
\usepackage{amssymb} 
\usepackage{amsmath}
\usepackage{subcaption}

\begin{document}

% Copyright
\setcopyright{acmcopyright}
%\setcopyright{acmlicensed}
%\setcopyright{rightsretained}
%\setcopyright{usgov}
%\setcopyright{usgovmixed}
%\setcopyright{cagov}
%\setcopyright{cagovmixed}

% DOI
\doi{10.475/123_4}

% ISBN
\isbn{123-4567-24-567/08/06}

%Conference
\conferenceinfo{PLDI '13}{June 16--19, 2013, Seattle, WA, USA}

\acmPrice{\$15.00}

%
% --- Author Metadata here ---
\conferenceinfo{WOODSTOCK}{'97 El Paso, Texas USA}
%\CopyrightYear{2007} % Allows default copyright year (20XX) to be over-ridden - IF NEED BE.
%\crdata{0-12345-67-8/90/01}  % Allows default copyright data (0-89791-88-6/97/05) to be over-ridden - IF NEED BE.
% --- End of Author Metadata ---

\title{
	A New Approach to Building the Interindustry Input--Output Table
	%A New Approach in Estimating the Interdependence of Industries
}
%\subtitle{[Extended Abstract]
%\titlenote{A full version of this paper is available as
%\textit{Author's Guide to Preparing ACM SIG Proceedings Using
%\LaTeX$2_\epsilon$\ and BibTeX} at

\numberofauthors{1} %  in this sample file, there are a *total*
\author{
	% 1st. author
	\alignauthor
	Ryohei Hisano \\ 
	\affaddr{Social ICT Research Center}\\
	\affaddr{Graduate School of Information Science and Technology}\\
	\affaddr{The University of Tokyo}\\
	\affaddr{Japan}\\
	\email{em072010@yahoo.co.jp}
	% 2nd. author
	%\alignauthor
	%G.K.M. Tobin\titlenote{The secretary disavows
	%any knowledge of this author's actions.}\\
	%       \affaddr{Institute for Clarity in Documentation}\\
	%       \affaddr{P.O. Box 1212}\\
	%       \affaddr{Dublin, Ohio 43017-6221}\\
	%       \email{webmaster@marysville-ohio.com}
}
% There's nothing stopping you putting the seventh, eighth, etc.
% author on the opening page (as the 'third row') but we ask,
% for aesthetic reasons that you place these 'additional authors'
% in the \additional authors block, viz.
%\additionalauthors{Additional authors: John Smith (The Th{\o}rv{\"a}ld Group,
%	email: {\texttt{jsmith@affiliation.org}}) and Julius P.~Kumquat
%	(The Kumquat Consortium, email: {\texttt{jpkumquat@consortium.net}}).}
% Just remember to make sure that the TOTAL number of authors
% is the number that will appear on the first page PLUS the
% number that will appear in the \additionalauthors section.

\maketitle
\begin{abstract}
	
We present a new approach to estimating the interdependence of industries in an economy by applying data science solutions.  By exploiting interfirm buyer--seller network data, we show that the problem of estimating the interdependence of industries is similar to the problem of uncovering the latent block structure in network science literature.  To estimate the underlying structure with greater accuracy, we propose an extension of the sparse block model that incorporates node textual information and an unbounded number of industries and interactions among them.  The latter task is accomplished by extending the well-known Chinese restaurant process to two dimensions.  Inference is based on collapsed Gibbs sampling, and the model is evaluated on both synthetic and real-world datasets.  We show that the proposed model improves in predictive accuracy and successfully provides a satisfactory solution to the motivated problem.  We also discuss issues that affect the future performance of this approach.
	
\end{abstract}



%\begin{CCSXML}
%<ccs2012>
%  <concept_id>10010520.10010553.10010562</concept_id>
%  <concept_desc>Computer systems organization~Embedded systems</concept_desc>
%  <concept_significance>500</concept_significance>
%  <concept_id>10010520.10010575.10010755</concept_id>
%  <concept_desc>Computer systems organization~Redundancy</concept_desc>
%  <concept_significance>300</concept_significance>
%  <concept_id>10010520.10010553.10010554</concept_id>
%  <concept_desc>Computer systems organization~Robotics</concept_desc>
%  <concept_significance>100</concept_significance>
%  <concept_id>10003033.10003083.10003095</concept_id>
%  <concept_desc>Networks~Network reliability</concept_desc>
%  <concept_significance>100</concept_significance>
%</ccs2012>  
%\end{CCSXML}

%\ccsdesc[500]{Computer systems organization~Embedded systems}


%
%  Use this command to print the description
%

% We no longer use \terms command


\section{Introduction}
\subsection{General Introduction}
% P1
The input--output table is a matrix that summarizes the interdependence of industries in an economy \cite{Miller2009}.  It is concerned with the activity of industries that buy goods produced by other industries for their own production.  Each row in the table represents the distribution of a producer's output to other industries and each column represents the composition of inputs required by a certain industry to produce its output.  The table is one of the fundamental statistics that describe the state of a macroeconomy assembled by governments worldwide and international organizations \cite{BEA,JIO,Timmer2015}.  It is used by academics, businessmen and government officials to capture the circular flow of transactions in an economy.

% P2
The basic methodology for assembling the input--output table was developed in the late 1930s.  Although various developments have been made, the basic methodology remains the same.  In this paper, we provide new methodology to solve the problem of summarizing the interdependence of industries in an economy.  The approach is, in essence, a dimensionality reduction technique that uses a graphical model to capture the dependence among multi-source datasets (i.e., the interfirm buyer--seller network and short textual information that summarizes firms' main business lines).  Although the motivation for this paper might be unfamiliar to the community, it shows the strength of using familiar machine learning techniques to answer real-world questions.  Furthermore, it creates an opportunity to explore new research challenges that concern economic networks.  

\subsection{Comaprison to the traditional approach}

% P3
The interdependence of industries is formed by firms' trade relationships of buying goods from other firms as an input to their own production.  Thus, discarding all the obstacles to gathering data, the ideal dataset for which we want to base the analytics starts at the firm level.  One example of this ideal dataset is summarized in Table~\ref{table1}.  This dataset contains all the information concerning {\it which firm bought what goods from which other firm to produce what goods}, together with the transaction date, price and volume of the purchased goods in an economy.  However, because of the limitations of current information gathering technology and privacy issues concerning firms' business strategies, currently, it is impossible to gather these ideal data.  Thus, to overcome this issue, the traditional input--output table is based on surveys and an interpretation of other primary and secondary economic data to gather information concerning {\it how much of what goods is bought to produce particular goods in an industry} (Table~\ref{table2}).  Together with the list of industries and goods supposed to be produced in an economy, several rounds of meetings are held by professionals and the coefficients in the input--output transaction table are determined.

% P4
An important factor to notice in the traditional approach is the absence of information concerning firms' trade relationships (columns one and four in Table~\ref{table1} and Table~\ref{table3}).  The main reason for this ignorance is due to data gathering issues.  However, at the current time, there are information providers that gather this type of data in various ways (i.e. questionnaire, press release, customs).  Moreover, information concerning firms, such as text describing firms' main business lines, web pages and detailed industry classifications, are also becoming increasingly available.  Under these emerging changes in the data environment, a new way to approach the problem is needed.

%I propose a new way to estimate the interdependence of industries in an economy.

\begin{table*}
	\centering
	\caption{Ideal data describing firms's trade relationships of buying goods from other firms as an input to their own production.} \label{table1}
	\begin{center}
		\begin{tabular}{|l|l|l|l|l|l|l|} \hline
			{\bf Buyer} &{\bf Purpose}  &{\bf Goods} &{\bf Seller} &{\bf Price} &{\bf Volume} &{\bf Date}\\
			\hline 
			Firm A &To produce car  &tyre &Firm B  &50 &2 &2015.9.1 \\
			Firm A &To produce car  &glass &Firm B  &40 &5 &2015.9.1 \\
			Firm A &To produce car  &aluminum &Firm C  &60 &3 &2015.9.1 \\
			Firm B &To produce tyre  &rubber &Firm D  &2 &20 &2015.9.1 \\
			\hline\end{tabular}
	\end{center}
\end{table*}

\begin{table*}
	\caption{Basic data used to assemble the input-output table} \label{table2}
	\begin{center}
		\begin{tabular}{|l|l|l|l|l|l|l|} \hline
			{\bf Purpose}  &{\bf Goods} &{\bf Total transaction}\\
			\hline 
			To produce car  &tyre &100 \\
			To produce car  &glass &200 \\
			To produce car  &aluminum &180 \\
			To produce tyre  &rubber &40 \\
			\hline
		\end{tabular}
	\end{center}
\end{table*}	

%\begin{table}[h] NARA MASANI KONO BASYO NI BYOUGA
\begin{table}
	\caption{Basic data used in this paper.  It summarizes the interfirm relation among firms.} \label{table3}
	\begin{center}
		\begin{tabular}{|l|l|l|l|l|l|l|} \hline
			{\bf Buyer} &{\bf Seller} \\
			\hline 
			Firm A &Firm B \\
			Firm A &Firm C   \\
			Firm B &Firm D \\
			\hline
		\end{tabular}
	\end{center}
\end{table}

%  This data plus the list of industries, list of goods produced in an economy and their correspondence table are used to determine the table.
%   This data plus short textual information summarizing each firms main business lines are used to determine the interdependence of industries in this paper

\subsection{Contributions}

% P6
In this paper, we first propose a model to estimate the interdependence of industries using interfirm buyer--seller network information and short textual information concerning firms' main business lines.  We then provide a quantitative comparison between the predictive performance of the model and previous related machine learning models using both synthetic and real datasets.  The quantitative experimental results are followed by a qualitative result that demonstrates how the proposed model could summarize the interdependence of industries in an economy.  We also provide direct comparison of the estimated interindustry structure from our approach to the input--output table.

% P5 
The proposed approach is distinct from the traditional approach in three ways.  First, it uses complementary information not used in the traditional approach (compare Tables~\ref{table2} and \ref{table3}).  This is not an argument regarding which method is better.  Instead, this paper asserts the usefulness of exploiting different parts of the ideal dataset not previously exploited ultimately being able to leverage more information than the traditional method.  Second, the unsupervised nature of the proposed approach makes it possible to estimate the interdependence of industries from the bottom up, thereby automatically estimating the industries involved in an economy.  This is in contrast to the traditional approach, where all the industries supposed to be operating in an economy have to be predefined, which might be problematic when trying to estimate an unconventional industry structure.  Third is extensibility.  The strength of the graphical modeling approach presented in this paper originates in its modularity.  Although we only use node textual information to leverage the understanding of the network in this paper, it is easy to extend the model to incorporate additional information, such as that summarized in Table~\ref{table1}.  Moreover, geographic information and multiplex relational information, in addition to interfirm buyer--seller information, could also be incorporated, which makes it possible to further exploit the various sources of information emerging from the changes in the data environment.

\subsection{Related works concerning our modelling}
% P7
The proposed model used to estimate the interdependence of industries in this paper is an extension of the sparse block (SB) model of Parkkinen et al. \cite{Parkkinen2009}, which jointly models network information and node textual information.  The joint modeling of network and textual information (i.e., interfirm buyer--seller relationships and a short line of text summarizing each firm's main business line) is an important step to effectively estimate the interdependence of industries.  Many authors \cite{Rossi2012,Newman2015} have considered the importance of using extra information to leverage our understanding of the network. We propose two models to enable the SB approach to jointly simulate network and node textual information.  Of these, one can be regarded as a direct counterpart of the relational topic model (RTM) \cite{Chang2009} and is related to the link-LDA method~\cite{Nallapati2008}.  These models combine latent Dirichlet allocation (LDA) \cite{Blei2003} with the mixed membership stochastic block (MMSB) model of Airoldi et al. \cite{Airoldi2008} and are widely used in the literature.

% P8
The advantage of using the SB model instead of its MMSB counterpart as the underlying generative process for network formation is its ability to exploit the SB structure of the network directly.  This view is shared with previous work dealing with the SB model \cite{Balasubramanyan2011,Peel2011,Gyenge2010,Ho2014,Peel2014}.  The fact that it is better to assume that most industry pairs have no interactions also originates from the specific dataset used in this paper.  In the dataset, each firm is requested to name up to five buyers of its products and suppliers of the intermediate goods used in its own production.  This scheme corresponds to the fixed rank nomination scheme in social network analysis (c.f. a friendship network) \cite{Hoff2013}.  For this scheme, all minor relationships\footnote{For instance, a firm manufacturing cars would not list a stationery store as one of its most important suppliers.} would be ignored in the network, which makes it possible to estimate only the major interdependence among industries.  The SB model enables us to exploit this SB structure more efficiently.% and with greater ease.  

%9
Additionally, the generating process of the SB model only models existing links (i.e. edge list) and ignores links that are not fomed.  Compared with the MMSB model, where both the existence and nonexistence of links are modeled, this saves many computations when the network structure is sparse.  This makes it suitable for large sparse graphs, such as the interfirm networks modelled in this paper.

% P11
The simultaneous estimation of the number of industries and active interactions among them is implemented by employing a two-dimensional extension of the Chinese restaurant process \cite{Pitman2006}.  Our motivation for extending the Chinese restaurant process to two dimensions originates from a need to model the SB structure among industries.  New link patterns among industries could either be generated from (i) new emerging industries or (ii) new link patterns emerging from already existing industries.  Both types of link formation (i.e., exogenous and combinatorial) are important to the innovation process of an interfirm buyer--seller network, and we use this as the prior process in our model.

%This extension is used to model the polyaidc relations involving industries and firms.  

% P12
There is one disadvantage of our extended model.  The Chinese restaurant process exhibits an important invariance property called exchangeability \cite{Gershman2012}, which makes inference via Markov chain Monte Carlo (MCMC) sampling straightforward. However, this is no longer the case in our two-dimensional extension.  This is a well-known issue when taking a sequential formulation (predictive distribution) approach to model the prior process \cite{Pitman2006,Wallach2010,Airoldi2014}.  One strategy is to follow previous works \cite{Miller2012,Airoldi2014} and directly process non-exchangeable priors by developing an appropriate inference methodology.  However, we show that the break in the invariance property is only slight, and a minor modification to the joint distribution suffices to recover the invariance property.  In the proposed model, we use the joint distribution with this approximation.  After recovering exchangeability via approximation, inference is performed using collapsed Gibbs sampling.  This is in line with previous Bayesian nonparametric models \cite{Griffiths2005,Kemp2006}.

\subsection{Organization of the paper}

% P13
The remainder of this paper is organized as follows:  In Section 2, we introduce the two basic models and illustrate their inference strategy.  In Section 3, we present the two-dimensional Chinese restaurant process  and demonstrate how the invariance property breaks down and is repaired in the joint distribution.  In Section 4, we combine the joint distribution, which is motivated by the two-dimensional Chinese restaurant process, with one of the two basic models described in Section 2. In Section 5, we evaluate the performance of the proposed methods.  In the final section, we discuss further related work and present the conclusion\footnote{Replication codes are available from the author's website.}.

%% 変に書く位なら書かない方がいいというのも事実。
%\section{Problem Statement} 
%An interfirm buyer--seller network $N$ is defined as 
%\begin{equation}
%N = (V,E)
%\end{equation}

%where V desribes a set of nodes and E describes a set of edges.  Instead of focusing on the adjacency matrix $A(N)$ where we define both existence and non-existence of links we focus on the edgelist $E$ which only records existing links.  For each nodes $v$ in $V$ we have a list of words $W_v$ describing the characteristics of each nodes.  This is denoted as the wordlist in this paper.  Using both edgelist and wordlist, the problem is to find a suitable interindustry matrix $W$ that best clusters each links in the interfirm buyer--seller network.  Evaluation of the clustering would be besed both on qualitative and quantitative results.

\section{Basic Models} 
\subsection{Sparse block model with text}

\begin{figure*}[h]
	\vspace{.2in}
	\begin{subfigure}{.35\textwidth}
		\includegraphics[width=0.6\linewidth]{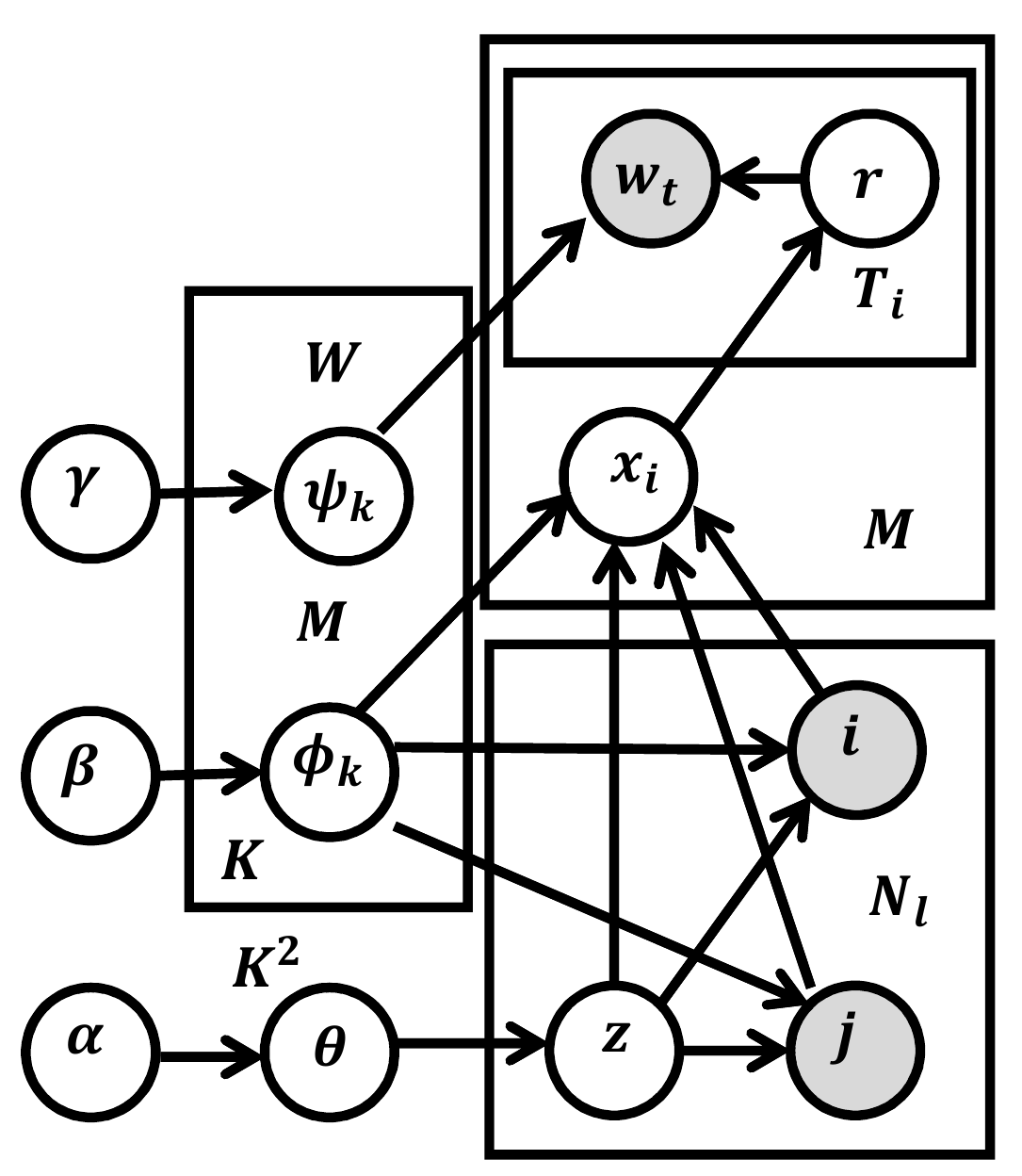}
		\caption{Sparse block model with text (SBT)}
		\label{fig1_1}
	\end{subfigure}%
	\begin{subfigure}{.35\textwidth}
		\includegraphics[width=0.6\linewidth]{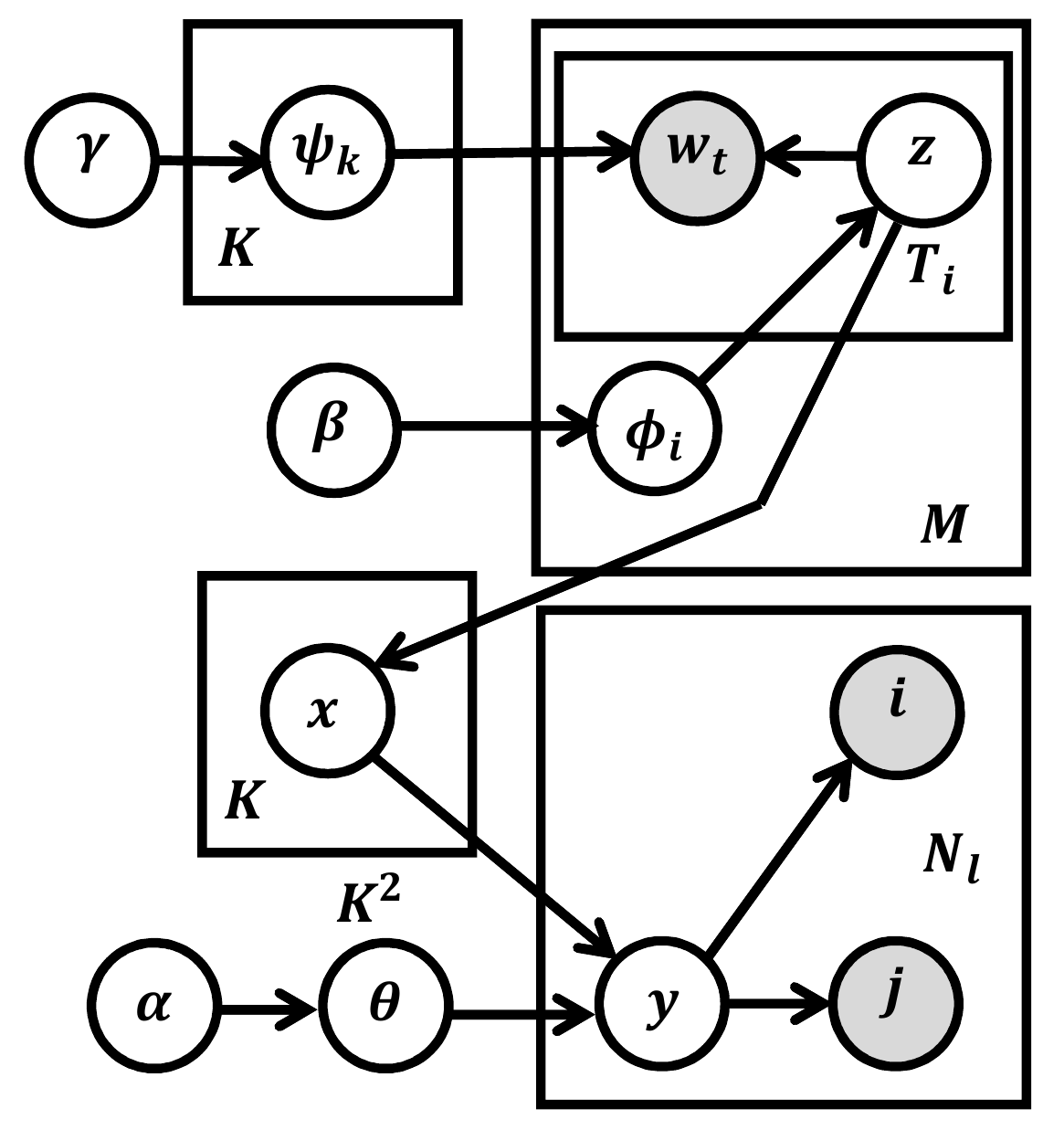}
		\caption{Reversed SBT}
		\label{fig1_2}
	\end{subfigure}
	\begin{subfigure}{.35\textwidth}
		\includegraphics[width=0.6\linewidth]{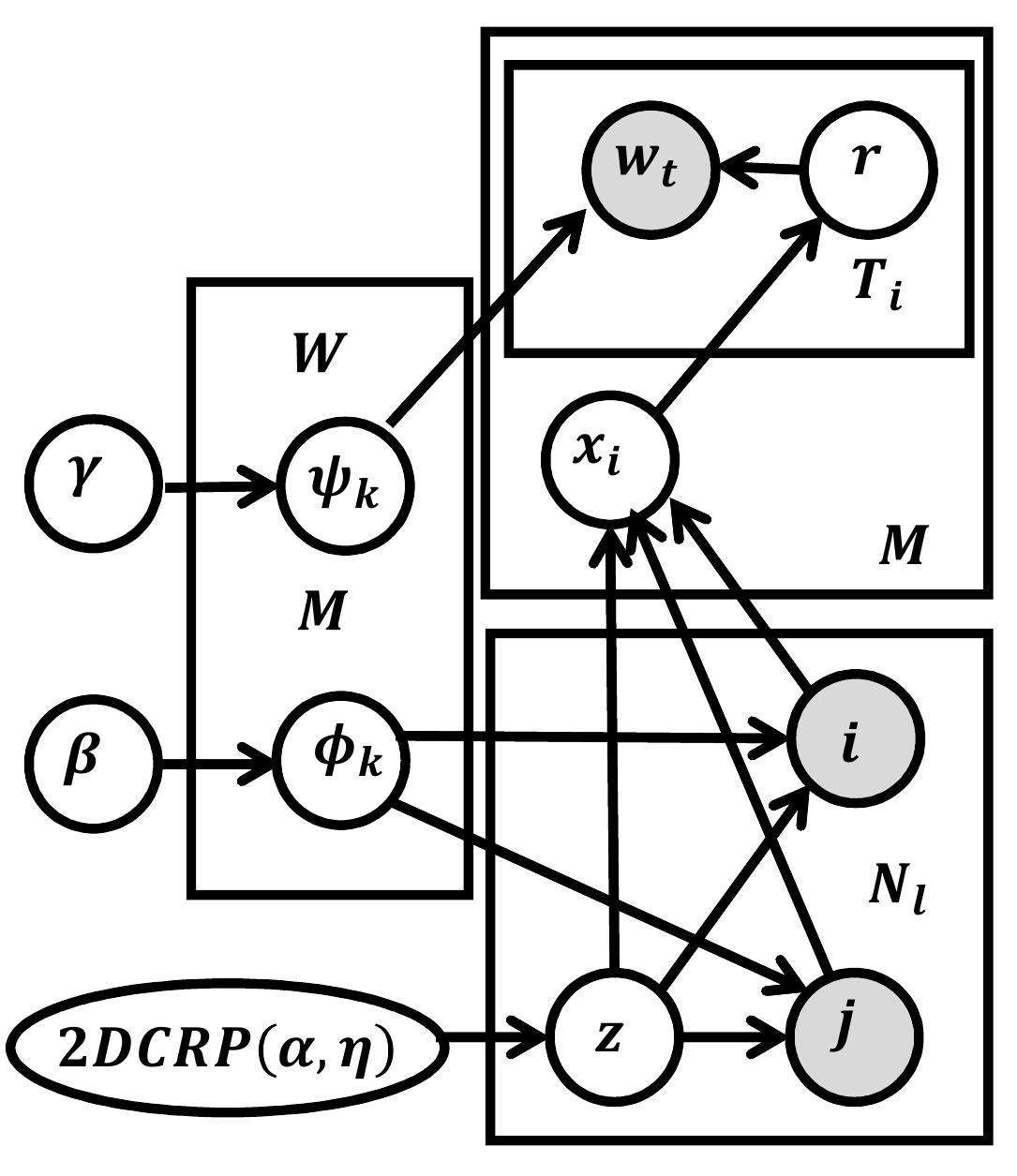}
		\caption{Infinite SBT}
		\label{fig1_3}
	\end{subfigure}
	\vspace{.2in}
	\caption{Plate diagram} 
	\label{fig:1s} 
\end{figure*}

To jointly model network and node textual information, we use a combination of the SB model~\cite{Parkkinen2009} and LDA~\cite{Blei2003}.  Figure~\ref{fig1_1} shows the plate diagram of the model.  The lower component corresponds to the SB model and the upper component represents LDA.  The generative process is comprised of the following two stages.

\noindent {\bf 1. Generate edge list:}

(1) Sample $\theta \sim Dirichlet(\alpha)$, where $\theta$ denotes the multinomial distribution over industry pair labels.  The dimension of this multinomial is $K^2$.

(2) Sample each $\phi_{k} \sim Dirichlet(\beta)$, where each $\phi_{k}$ denotes the multinomial distribution over firms.  There are $M$ firms in the network, and from $\phi_{k}$, we can sample a firm from industry number $k$.

(3) For each edge, first sample an industry pair $(z_{1},z_{2}) \sim Multinomial(\theta)$ and then sample firms from each industry via $i \sim Multinomial(\phi_{z_{1}})$ and $j ~ \sim Multinomial(\phi_{z_{2}})$.  This completes the generation of an edge list.  There are $N_{l}$ edges in total.

\noindent {\bf 2. Generate word list:}

(1) Sample each $\psi_{k} \sim Dirichlet(\gamma)$, where each $\psi_{k}$ denotes the multinomial topic distributions.  There are $W$ words in the vocabulary list. 

(2) For each firm $i$ in the network, consider the distribution of an industry number involving $i$ as either a sender (i.e., $z_{1}$) or receiver (i.e., $z_{2}$).  The distribution of industry numbers for a given firm $i$ in the edge list is denoted by $x_{i}$.  Note that this could be approximated as
$x_{ik} \sim \phi_{ki}\frac{M\beta+\Sigma_{j=1}^{M}q_{z_1=k,j}+q_{z_2=k,j}}{K\beta+\Sigma_{l=1}^{K}q_{z_{1}=l,i}+q_{z_{2}=l,i}}$, where $q_{z_{1}=k,j}$ counts the number of edges in which node $j$'s industry number as a sender is $k$, $q_{z_{2}=k,j}$ counts the number of edges in which node $j$'s industry number as a receiver is $k$, because  $\phi_{ki}$ and $x_{ik}$ share the same numerator.  To elaborate on this point, $\phi_{ki}$ could be calculated as the number of times $i$ is labeled industry number $k$ as either a sender or receiver divided by the number of times industry number $k$ appeared in the edge list, whereas $x_{ik}$ could be calculated by the number of times $i$ is labeled industry number $k$ as either a sender or receiver divided by the number of times $i$ appeared in the edge list.  Hence, the only difference between the two is the denominator.

(3) There are $T_{i}$ words in firm $i$'s short text. For each word position $t$ in firm $i$'s text, sample $r_{t} \sim Multinomial(x_{i})$.

(4) Finally, sample each word from the topic distribution $\psi_{r_{t}}$.  There are $\Sigma_{i=1}^{M}T_{i}$ words in total in the entire corpus.

Note that creating an edge directly from $\phi_{k}$ to $x_{i}$ risks the separation of the multinomials into those that explain the edge list and the other word list.  This separation was also noted by Chang and Blei~\cite{Chang2009} as a criticism of the link-LDA model~\cite{Nallapati2008}.  Following previous work~\cite{Chang2009}, we force parameter sharing by linking the sampled $z$ directly to $x_{i}$.

There are two previous works that are extremely close to our model.  One is the block-LDA model of Balasubramanyan and Cohen~\cite{Balasubramanyan2011}, which also combines the SB model with LDA.  However, for block-LDA, the focus is to model links between entities in the documents instead of the link between documents.  In the application we consider (i.e., interfirm buyer--seller networks), node textual information is provided for each firm. Thus, we avoid using the block-LDA model.  Another work that is similar to our model concerns topological feature classification~\cite{Peel2011}.  The difference between this model and the proposed model is that each node has a class label instead of textual information.  For instance, if we used industrial classification rather than textual information, the approach of Peel \cite{Peel2011} might have been more appropriate.  However, because we do not wish to set the number of roles as the number of prescribed industry classifications and use more information than simply the industry classification by exploiting textual information about each firm's business, we also avoid using the topological feature-based classification.  The likelihood of the model is 

\begin{equation}
	\begin{split}
		p(L,Z,W,R,\psi,\phi,\theta|\alpha,\beta,\gamma) = D\prod_{z=1}^{K^2}\theta^{n_z+\alpha-1} \\
		\prod_{k=1}^{K}\prod_{i=1}^{M}\phi_{ki}^{q_{z_1=k,i}+q_{z_2=k,i}+\beta-1}  \prod_{k=1}^{K}\prod_{w=1}^{W}\psi_{kw}^{\gamma-1+\Sigma_{i=1}^{M}\Sigma_{t=1}^{T_i}r_{it}=k,w} \\
		\prod_{i=1}^{M}\prod_{k=1}^{K}(\phi_{ki}\frac{M\beta+\Sigma_{j=1}^{M}q_{z_1=k,j}+q_{z_2=k,j}}{K\beta+\Sigma_{l=1}^{K}q_{z_{1}=l,i}+q_{z_{2}=l,i}})^{r_{k}^{i}}
	\end{split}
\end{equation}

\noindent
where $M$ denotes the number of nodes, $K$ denotes the number of industries (i.e., topics), $T_i$ denotes the number of words in the textual information for node $i$, $W$ denotes the number of unique words in the node textual information, $D$ is a normalizing constant, $n_z$ denotes the number of times block pair $z$ has been sampled, $q_{z_{1}=k,i}$ counts the number of edges in which node $i$'s industry number as a sender is $k$, $q_{z_{2}=k,i}$ counts the number of edges in which node $i$'s industry number as a receiver is $k$, $r_{it}=k,w$ denotes the number of word positions of node $i$ that have word $w$ and topic number $k$, $r_{k}^{i}:=\Sigma_{w=1}^{W}\Sigma_{t=1}^{T_i}(r_{it}=k,w)$ and the remaining quantities denote multinomial distributions and hyperparameters. Note that the last term is an approximation that uses the fact that $\phi_{ki}=\frac{f(k,i)}{\Sigma_{i}f(k,i)}$ and $x_{ik}=\frac{f(k,i)}{\Sigma_{k}f(k,i)}$ share the same numerator, as already noted above.

The collapsed Gibbs sampler for each edge (i.e., $p(z_{0}|.)$ where $z_0$ denotes the industry pair of the link we are sampling) and word position (i.e., $p(r_{0}|.)$ where $r_{0}$ denotes the industry of the word position we are sampling) could be derived by taking exactly the same step as that of LDA~\cite{Griffiths2004}.  We omit the derivation to save space.  %For ease of replication, the sampler is shown in the Appendix.

\subsection{Reversed sparse block model with node textual information}

The generating process described above can be reversed. First, the word list is generated and then the inferred multinomial distributions are used to generate the edge list. Figure~\ref{fig1_2} shows the plate diagram of this model.  The generative process is as follows:

\noindent {\bf 1. Generate word list:}

(1) Sample each $\psi_{k} \sim Dirichlet(\gamma)$, where each $\psi_{k}$ denotes the multinomial topic distributions . There are $W$ words in the vocabulary list. 

(2) For each firm $i$ in the network, sample its topic proportion as $\phi_{i} \sim Dirichlet(\gamma)$.

(3) There are $T_{i}$ words in firm $i$'s short text. For each word position $t$ in firm $i$'s text, sample each topic as $z_{t} \sim \phi_{i}$.

(4) Sample each word $w_{t}$ from topic distribution $\phi_{z_{t}}$.  There are $\Sigma_{i=1}^{M}T_{i}$ words in total in the entire corpus.

\noindent {\bf 2. Generate edge list:}

(1) Sample $\theta \sim Dirichlet(\alpha)$, where $\theta$ denotes the multinomial distribution over industry pair labels. The dimension of this multinomial is $K^2$.

(2) For all firms in the network, consider the distribution of the topic number (industry number) for words in each firm's short text.  The distribution of firms for a given industry $k$ is denoted by $x_{k}$.  This could be approximated as $x_{ki} \sim \phi_{ik}\frac{K\beta+\Sigma_{k=1}^{K}n_{z}(i,k)}{M\beta+\Sigma_{i=1}^{M}n_{z}(i,k)}$ because, as before, $\phi_{ik}$ and $x_{ki}$ share the same numerator.  There are $M$ firms in the network.

(3) For each edge, first sample the industry pair $(y_{1},y_{2}) \sim Multinomial(\theta)$ and then sample firms from each industry using $i \sim Multinomial(x_{y_{1}})$ and $j ~ \sim Multinomial(x_{y_{2}})$.  This completes the generation of the edge list.

Although this model is essentially the same as that described above, we also measure the predictive performance of this model because it closely resembles the RTM in which we start with LDA and generate the network structure using MMSB \cite{Chang2009}.  The sampler of this model could be derived in a similar manner. % and is shown in the Appendix.

\section{Two-dimensional Chinese restaurant process} 

To enable the aforementioned model to process a potentially infinite number of industries and the interdependence among them, we introduce a new prior distribution by extending the Chinese restaurant process.  A schematic figure describing the process is shown in figure 2.  As with other Bayesian nonparametric models, we derived the distribution via defining a sequential process.  There are other ways to define exactly the same joint distribution as that of Equation (7).  However, the following illustration, which adheres to a sequential formulation, makes it clear how the process distinguishes between the creation of new industry pairs involving new industries (i.e. figure 2d) and the creation of new industry pairs using a combination of industries that already exist (i.e. figure 2c).  Both types of innovation are well known to exist and thus, we derive the distribution from the sequential formulation.

% 2DCRP
\begin{figure*}
	\begin{minipage}[b]{0.24\linewidth}
		\centering
		\includegraphics[keepaspectratio, scale=0.8]
		{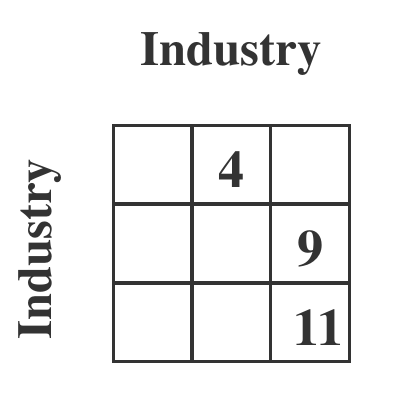}
		\subcaption{At time $t-1$.}
		\label{fig_A}
	\end{minipage}
	\begin{minipage}[b]{0.24\linewidth}
		\centering
		\includegraphics[keepaspectratio, scale=0.8]
		{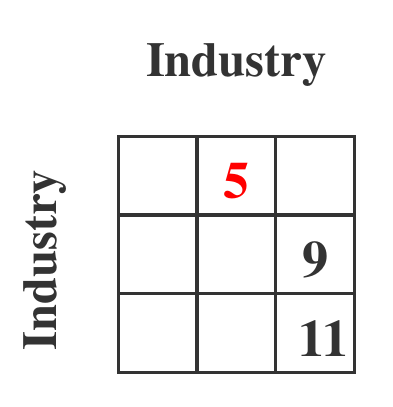}
		\subcaption{At time $t$.  Existing industry pair gains another link.}
		\label{fig_B}
	\end{minipage}
	\begin{minipage}[b]{0.24\linewidth}
		\centering
		\includegraphics[keepaspectratio, scale=0.8]
		{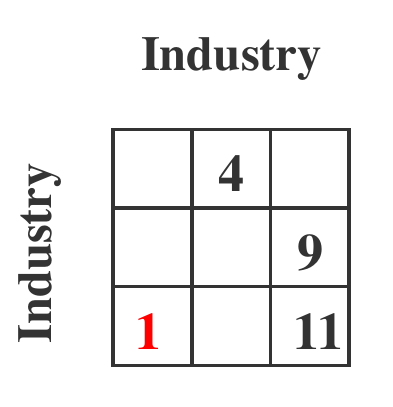}
		\subcaption{At time $t$.  New industry pair created.  The rate of occurrence is controlled by $\alpha$.}
		\label{fig_C}
	\end{minipage}
	\begin{minipage}[b]{0.24\linewidth}
		\centering
		\includegraphics[keepaspectratio, scale=0.7]
		{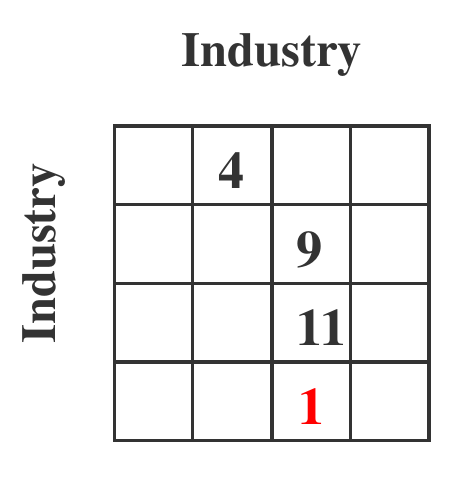}
		\subcaption{At time $t$.  New indusry and new industry pair created.  The rate is controlled by $\eta$.}
		\label{fig_D}
	\end{minipage}
	\caption{Schematic figure describing the 2DCRP.  (a) describes the state of the interindustry network at time $t-1$.  (b-d) descibes the three type of cases which might follow in the next time step.}
	\label{fig2D}
\end{figure*}

At the start of the enterprise, suppose that a pair of firms both in industry A establishes a trade relationship.  The second pair of firms to arrive in the economy first tries to create a new industry B with probability $\eta$.  If the pair succeeds in creating the new industry, the pair could both belong to industry B or the link could be classified as a combination involving A (e.g., AB or BA).  The ordering of pairs is important. If AB is chosen, this implies that the firm in industry A sells goods to the firm in industry B, and vice versa.  If the second pair fails to create a relationship involving a new industry, this pair could just follow the first pair and connect firms among industry A.  Suppose now that the second pair succeeds in creating a new industry and links firms between industries AB.  Under this scenario, the third pair to arrive in the economy now has a third option.  This pair could either create a new link involving a new industry C or create a new link among industry pairs that nobody has created before (e.g., BB or BA), with some probability governed by the parameter $\alpha$, or this pair could follow the first and second pairs and create links among industries according to the industry pairs' popularity.  The three distinct behavior is summarized in figure 2.

To illustrate the process in more detail with an example, consider the following process. 

\begin{enumerate}
	\item A trade relationship is established between two firms both in industry A.
	\item The second pair of firms succeeds in creating a new industry and links are established among industries AB (i.e., firms in A sell to firms in B).	
	\item The third pair of firms decides to follow the first pair and forms a new link among firms in industry A.
	\item The fourth pair of firms decides to follow the first pair and forms a new link among firms in industry A.
	\item The fifth pair of firms determines that one could create a new link among firms both in industry B and decides to create a new link.
	\item The sixth pair of firms succeeds in creating a new link involving a new industry C that connects firms among industries CA (i.e., firms in C sell to firms in A).
	\item The seventh pair of firms decides to follow the second pair and forms a new link among firms among industries AB (i.e., firms in A sell to firms in B).
\end{enumerate}

From the illustration above, the joint distribution of the above process can be written as

\begin{equation}
	\begin{split}
		p(z_{1:10}|\alpha,\eta)= \frac{\eta}{1+\eta}\frac{1}{3}\frac{2}{2+\eta}\frac{\alpha+1}{4\alpha+2}\frac{3}{3+\eta}\frac{\alpha+2}{4\alpha+3} \\
		\frac{4}{4+\eta}\frac{\alpha}{4\alpha+4}\frac{\eta}{5+\eta}\frac{1}{5}\frac{6}{6+\eta}\frac{\alpha+1}{9\alpha+6}\frac{7}{7+\eta}\frac{\alpha+3}{9\alpha+7}
		%\\ \frac{8}{8+\eta}\frac{\alpha+2}{9\alpha+8}\frac{9}{9+\eta}\frac{\alpha}{9\alpha+9}.
	\end{split}
\end{equation}

Note that $\eta$ controls the probability of creating a new industry while $\alpha$ controls the probability of creating new edges (i.e. Figure 2).  It is important to note that exchangeability is broken because the denominator of the 4, 6, 8, 12,... terms, namely

% depends on the timing when the new element was created.  However rearranging terms, the contribution of each components to the joint likelihood is either 
\begin{equation}
	\begin{split}
		\frac{2}{4\alpha+2}\frac{3}{4\alpha+3}\frac{4}{4\alpha+4}\frac{6}{9\alpha+6}\frac{7}{9\alpha+7}%\frac{8}{9\alpha+8}\frac{9}{9\alpha+9}
	\end{split}
\end{equation}

\noindent depends on when the new industry was created.  To summarize, based on the above sequential formulation, the exact timing for the creation of a new industry influences the probability of industry pairs using already existing industries to emerge.  Rearranging the above terms, the contribution of each component to the joint likelihood is given by

\begin{equation}
	\begin{split}
		p(z_{1:N_l}=z|\alpha,\eta)=\frac{(I_{z,1}-1)(I_{z,2}-1)\cdots(I_{z,N_z}-1)}{(I_{z,1}-1+\eta)\cdots(I_{z,N_z}-1+\eta)}\\
		\frac{\alpha}{I_{z,1}-1+K_{A(z,1)}^{2}\alpha}\frac{n_z-1+\alpha}{I_{z,2}-1+K_{A(z,2)}^{2}\alpha}\cdots \\
		\frac{1+\alpha}{I_{z,2}-1+K_{A(z,N_z)}^{2}\alpha}
	\end{split}
\end{equation}

\noindent
for industry pairs that were first produced by combining existing industries and

\begin{equation}
	\begin{split}
		p(z_{1:N_l}=z|\alpha,\eta)=\frac{\eta(I_{z,2}-1)\cdots(I_{z,N_z}-1)}{(I_{z,1}-1+\eta)\cdots(I_{z,N_z}-1+\eta)}\\ \frac{(n_z-1+\alpha)\cdots(1+\alpha)}{(I_{z,2}-1+K_{A(z,2)}^{2}\alpha)\cdots(I_{z,2}-1+K_{A(z,N_z)}^{2}\alpha)}\\
		\frac{1}{K_{z}^{2}-(K_{z}-1)^{2}}
	\end{split}
\end{equation}

\noindent
for industry pairs that were first produced by adding a new industry to the economy, where $I_{z,i}$ denotes the identifier of the $i$th entrepreneur who first created a link using industry $z$.  Note that when $\eta$ and $\alpha$ are sufficiently close to 0, $K_{A(z,N_z)}^{2}\alpha$ increases slowly compared with $I_{z,N_z}-1$, which makes it possible to approximate

\begin{equation}
	\frac{(I_{z,2}-1)\cdots(I_{z,N_z}-1)}{(I_{z,2}-1+K_{A(z,2)}^{2}\alpha)\cdots(I_{z,2}-1+K_{A(z,N_z)}^{2}\alpha)}
\end{equation}
\noindent
simply as $1$.  With this approximation, we determine the joint distribution of the process for $N_l$ links, which can be written as

\begin{equation}
	\begin{split}
		p(z_{1:N_{l}}|\alpha,\eta)=D\frac{\Gamma(\eta)\prod_{z=1}^{K^2}\Gamma(n_z+\alpha)}{\Gamma(N_l+\eta)\Gamma(\alpha)^{K^2}}(\frac{\eta}{\alpha})^{K}\\
		\prod_{k=1}^{K}\frac{1}{k^2-(k-1)^2},
	\end{split}
\end{equation}

\noindent
where $D$ is a normalizing constant, $K$ denotes the number of industries (i.e., topics) and $n_z$ denotes the number of times block pair $z$ has been sampled.

\section{Infinite sparse block with node textual information} 
We use the joint distribution derived in the previous section as the prior distribution of the proposed model.  The sampler for the infinite SB with node textual information (InfSBT) model is as follows:  For a particular edge, we sample from

\begin{equation}
	\begin{split}
		p(z_{0}|.) \propto 
		\left(
		\begin{array}{c}
			\frac{n_{z_0}'+\alpha}{N_{l}-1+\eta}  \\
			\frac{\alpha}{N_{l}-1+\eta}  \\
			\frac{\eta}{N_{l}-1+\eta}\frac{1}{(K+1)^2-K^2}
		\end{array}
		\right) ...,
	\end{split}
\end{equation}

\noindent
where $z_0$ denotes the one pair of nodes which we are sampling and the terms in addition to the sampler of the industry pairs are exactly the same as those of the SBT model and hence, are omitted here to save space. The first branch corresponds to the case where an existing combination is sampled according to its popularity, the second branch corresponds to the case where a new combination is sampled using already existing industry pairs and the final branch corresponds to the case where a new combination is sampled using a new industry as either the sender or receiver.  The sampler for a particular word-topic pair is the same as mentioned previously, thus it is not reproduced here.

\section{Results} 
\subsection{Synthetic data}

To demonstrate the performance of our model, we first use a synthetic dataset.  The network consists of 70 nodes and 249 edges. The edges are randomly sampled from one of the 20 active interactions among 16 industries.  The edge list is accompanied by a word list.  Each node is associated with 0--12 words from the topic distributions (industries) in which they are involved.  The word list is almost the same length as the edge list and consists of 230 words in total.  Our goal is to simultaneously estimate the number of industries, underlying block structure and topic distribution from the randomly permuted colorless version of the network.

\textbf{Evaluation of the estimated block structure} We first examine how well the proposed model is able to determine the true block structure governing the synthetic data.  For comparison, we also report the results given by the SB model~\cite{Parkkinen2009}.  For models other than the infinite version, we set the number of industries to 16, which is the true number of blocks used to generate the dataset.  Although the hyperparameters could be estimated using a maximum a posteriori estimate or full Bayesian approach, we set them to 0.05 for ease of computation.  For all the finite models, we perform 50,000 iterations, with the final realization used for evaluation.  The performance is compared using two measures: variation of information (VI)~\cite{Meila2007} and the absolute error (AE) between the true and estimated block structures.  Every measure requires the ground truth network, and the latter measure requires the additional constraint that the number of industries is the same as the true number of blocks.  We report these two measures for both the edge list and word list.

%\begin{table}[h]
\begin{table}
	\caption{Evaluation of the estimated block structure.} \label{table4}
	\begin{center}
		\begin{tabular}{|l|l|l|l|l|l|l|}\hline
			{\bf Model} &{\bf NetAE}  &{\bf NetVI} &{\bf TopicAE} &{\bf TopicVI}\\
			\hline 
			SB &268  &2.91 &NA  &NA \\
			SBT &140 &\textbf{1.53} &\textbf{92} &\textbf{1.12} \\
			RevSBT &246  &1.92 &94  &1.24 \\
			InfSBT &\textbf{138}  &1.64 &\textbf{92}  &1.16 \\
			\hline
		\end{tabular}
	\end{center}
\end{table}

Table~\ref{table4} reports the results.  ``Net'' represents the estimation performance for the underlying block structure and ``Topic'' represents that for topic distributions. AE represents the absolute error and VI represents the variation of information.  It shows that the use of additional textual information enables the proposed models to significantly outperform the SB model.  Moreover, there is little difference between the performance of the SB model with node textual information and its reverse counterpart for estimating the topic distributions.  However, for estimating the block structure, the SB model with node textual information is clearly superior to its reverse version.  Furthermore, the infinite version performs almost as well as its finite counterpart.

\textbf{Predictive performance} Next, we compare the predictive performance of the model for both the edge list and word list.  Similarly to other probabilistic models, InfSBT defines a probability distribution over the given data.  However, compared with MMSB or RTM, which explicitly model the existence and nonexistence of a link between two nodes, the SB model and InfSBT only model the probability of a certain edge list occurring\footnote{Hence, in the sparse block-type models, the same edge could occur more than once.  This feature might be problematic when the network is dense, but this is not the case for the datasets used in this paper.}.  Therefore, simply dividing the edge list into training and test sets would create a test set that only consists of one label (i.e. existence of a link because the data is an edge list), which makes it difficult to use traditional measures such as area under the reciver operator curve (AUC), which requires both 0 (i.e. nonexistence of a link) and 1 (i.e. existence of a link) labels in the test set.  Thus, to evaluate the predictive performance of these two model types, we first define a score function to compare the models without adjusting the test set data.

For all possible links given a set of nodes, we evaluate the probability of a link being connected (for MMSB-type models) or the event probability that a link is generated from all possible links (for SB-type models).  We then rank each possible link in decreasing order of probability.  The average rank in the test edge lists is used as the evaluation score.

The predictive performance of the unseen edge list is evaluated by dividing the edge list into 10 sets.  For each set, we train the MMSB model, RTM, SB model and InfSBT model.  The number of groups is set to 16 in all models except InfSBT.  For MMSB and RTM, we use the codes provided by the author of RTM \cite{Chang2009,Chang2010}.  We also compare the performance with the null model in which the probability of each link is randomly ordered.  The first row of Table~\ref{table5} reports the average score from the 10 sets.  It is apparent that, without additional sparsity constraints, the SB model outperforms the MMSB model and RTM.

%\begin{table}[h]
\begin{table}
	\caption{Average score for edge list prediction} \label{table5}
	\begin{center}
		\hspace*{-0.3cm}
		\begin{tabular}{|l|l|l|l|l|l|}\hline
			{\bf Data} &{\bf Null}  &{\bf MMSB} &{\bf RT} &{\bf SB} &{\bf InfSBT}\\
			\hline 
			Synthetic &1827 &1773 &1623 &1111 &\textbf{794.6} \\
			Real &24572 &8228 &18328 &7934 &\textbf{7189} \\
			\hline
		\end{tabular}
	\end{center}
\end{table}

To compare the performance of the model using traditional measures (i.e., AUC), for each training and test set pair, we modify the test dataset in the following way.  For each test dataset, we randomly add 500 node pairs without links.  For this modified dataset, we calculate the ROC curve using each model and calculate AUC for each of the models using a different split of the dataset and calculate its mean value.  Table~\ref{table6} reports the result.  As before, we observe that InfSBT outperforms other methods.

%\begin{table}[h]
\begin{table}
	\centering
	\caption{Average AUC score for link prediction} \label{table6}
	\begin{center}
		\hspace*{-0.3cm}
		\begin{tabular}{|l|l|l|l|l|l|}\hline
			{\bf Data}  &{\bf MMSB} &{\bf RT} &{\bf SB} &{\bf InfSBT}\\
			\hline 
			Synthetic  &0.557 &0.584 &0.692 &\textbf{0.804} \\
			Real       &0.614 &0.544 &0.817 &\textbf{0.839} \\
			\hline
		\end{tabular}
	\end{center}
\end{table}

Word prediction is performed in the usual way.  We first randomly divide the word list into a training set (90\%) and test set (10\%), and ensure that each node has at least one word in the training set.  We compare the predictive performance of the test set in terms of its predictive log-likelihood.  For this task, we compare the proposed model to LDA and RTM.  The first row of Table~\ref{table7} shows the results.  It shows that taking network information into account results in better predictive performance in terms of the unseen words.

%\begin{table}[h]
\begin{table}[t]
	\caption{Predictive log-likelihood of the test word list} \label{table7}
	\begin{center}
		\begin{tabular}{|l|l|l|l|}\hline
			{\bf Data}  &{\bf LDA} &{\bf RTM} &{\bf InfSBT}\\
			\hline 
			Synthetic        &-81.17 &-88.58 &\textbf{-47.61} \\
			Real firm        &-638.41 &-661.83 &\textbf{-603.64} \\
			\hline
		\end{tabular}
	\end{center}
\end{table}

\subsection{Interfirm buyer--seller network}

We also apply the InfSBT model to a real-world interfirm buyer--seller network.  The network data are obtained from a data provider who collects interfirm buyer--seller information about Japanese firms\footnote{Tokyo Shoko Research Ltd.}.  In the dataset, each firm is requested to name up to five buyers of their products and suppliers of the intermediate goods used in their own production.  This scheme corresponds to the fixed rank nomination scheme in social network analysis (c.f. a friendship network).  The fixed rank nomination scheme has both advantages and disadvantages.  The advantage is that we can focus on the major relationships operating in an economy.  The disadvantage is that minor interdependence of industries is prone to be omitted from the estimate.

We use a subset of these data from the accounting year 2012 and focus on medium-sized firms and their surroundings.  The resulting network includes 222 firms.  For each firm, we obtain node textual information, written in Japanese, that describes its main business lines.  We first parse this information using a morphological analysis technique~\cite{Kudo2004} and select a word list containing only nouns.  We also delete several stop words.

To be more precise about the node textual information, it is a short text that describes the main business lines of a company.  If this information is well organized with a harmonized code, group labels might be easier to determine, but this is rarely the case.  Moreover, there are firms that conduct unusual business that cannot be fully captured using only industrial classification (group labels).  Furthermore, although we did not use it in this paper, there is additional textual information that provides an overview of the company.  Further research should include this.  Thus, in the paper, we choose to model textual information rather than group labels.

\begin{figure*}
	\centering
	\includegraphics[width=0.7\linewidth]{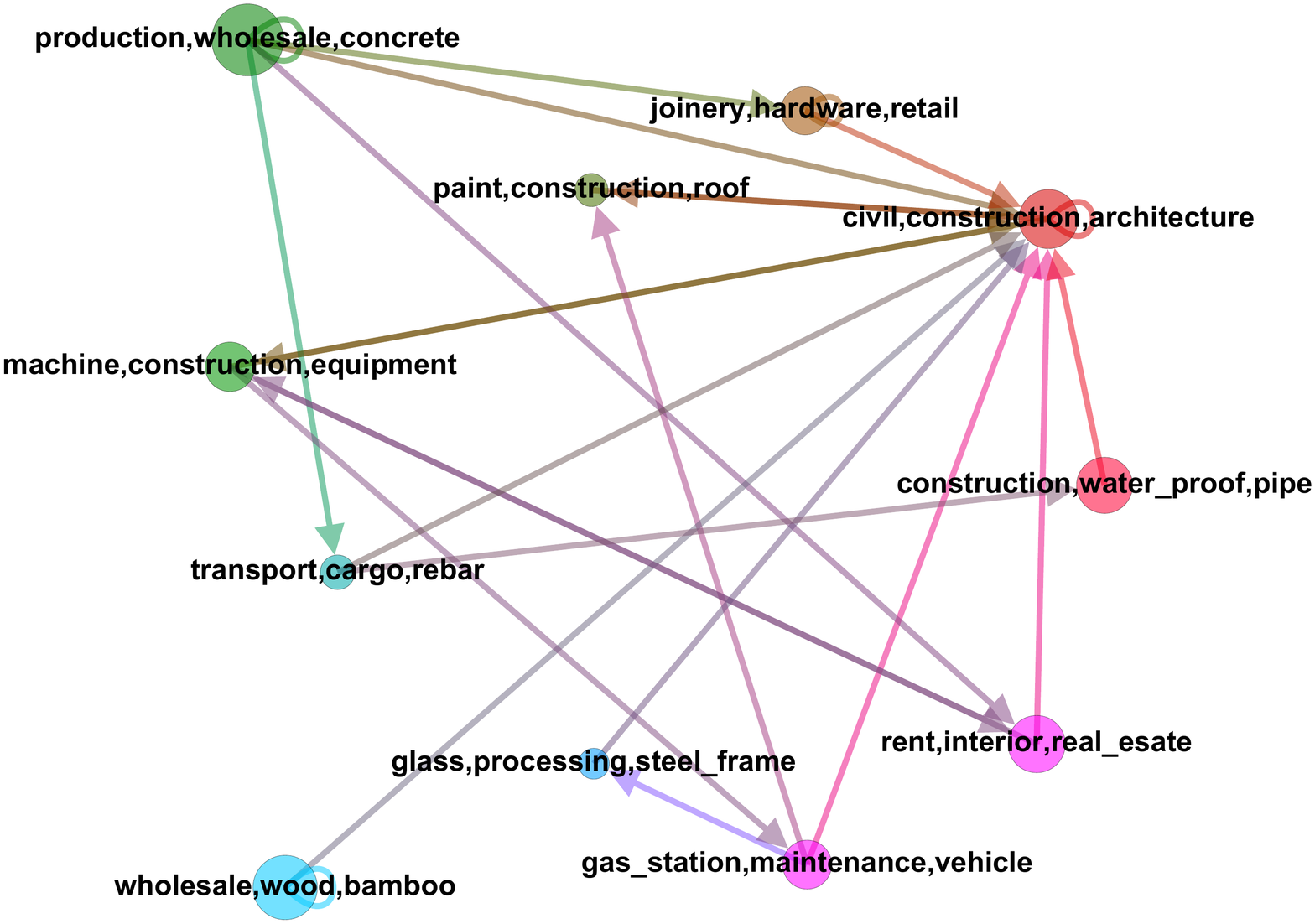}
	\caption{Network block structure estimated using InfSBT.  Node size is adjusted by their number of links.}
	\label{fig4_2}
\end{figure*}

MCMC is run for 2,000,000 iterations and we use the final realization for our network visualization.  The color and position of each industry are determined using polar coordinates.  Figure~\ref{fig4_2} reports the underlying block structure estimated from our model and shows that our proposed method successfully determines the underlying block structure.  There is a general construction business in the upper-right part of the network that purchases goods from other businesses (e.g., hardware, concrete, machinery, wood, glass, interior, cargo services and pipes).  There is another input--output relationship that describes the wholesale business centered on concrete that is sold to joinery and interior businesses.

Compared with the input--output table, the estimated industry structure is sparser.  In the input--output table, there are basic industries, such as electricity, real estate, banking and office supplies, that supply all other industries, which makes the interdependence of industries rather dense \cite{Miller2009}.  This is expected because of the use of network data with a fixed rank nomination scheme, where each firm only nominates up to five major buyers and sellers.  This might not be a problem when one is interested in the major relationships and omits minor relationships (e.g., firms buying stationery goods from a stationery store).  However, in the traditional input--output table, one focuses more on {\it what goods is bought to produce a particular goods}, which makes it possible to take these types of minor relationships into account.  For our approach to be able to incorporate these minor relationships, more data close to the ideal data in Table~\ref{table1} is required.  Despite the fact, the success of estimating a meaningful structure from the approach creates the opportunity for further extension that uses more elaborate models that could exploit multi-source datasets.  Further direct comparison between our estimate and the input--output table would be provided in the next subsection.

%.  Labels are the top three words from a topic distribution.

\textbf{Predictive performance} We also compare the predictive performance for these data.  The bottom row of Tables \ref{table5} and \ref{table6} reports the results for edge list prediction and shows that the proposed model outperforms all other models.  One reason for the poor performance of RTM may be that the implementation provided by J. Chang and M. J. Chang \cite{Chang2010} only models block diagonal interactions (i.e., assortative communities), whereas the networks studied in this paper display a more disassortative nature~\cite{Moore2011}.

The word prediction performance was also evaluated for this dataset.  The bottom row of Table 7 reports the results, which show that, as with the synthetic dataset, InfSBT outperformed all the other models.

\subsection{Comaprison to the traditional input--output table}

% P1
Compared to the traditional approach, the main feature of the presented approach lies in being able to directly assign each interfirm link an interindustry pair.  Thus, it is possible to look up the list of firms classified in an industry or list of links classified in an interindustry pair in a direct manner.  By design, the traditional approach ignores the interfirm relations and could not perform this direct lookup making the proposed approach complementary.

% P2
Putting aside this microlevel aspect, how similar or different are the interindusty structure estimated by the two approaches?  We present further qualitative comparison to answer this question.

Figure 4a shows the regional input--output table of the Nemuro-Kushiro area.  The Nemuro-Kushiro area is located at the east part of Hokkaido, Japan.  Only top edges sorted by their strength of relations are shown to aid visual comaparison.  We see that fishery, crop farming and construction is a major industry in this area showing strong linkage among the industry pairs.  Figure 4b shows the output using the presented approach\footnote{We selected 3,514 firms located in this area from the same data set used throughout the paper.}.  Four things are worth mentioning.  First of all, suspecting the topic words, we could confirm that fishery, food and construction are major industries in this area.  It is worth emphasizing that this result was achieved using completely different data set (as summarized in Table 2 and Table 3) and methodology compared to the traditional approach.  Secondly, we see that the network structure of the construction sector is more complicated than figure 4a suggests having multiple grups classified as a construction sector.  This is not surprising because in Japan there are a lot of small sized firms involved in the construction business connected in a complicated manner.  By exploiting the micro level interfirm network data, we are able to separate the construction sector into distinct groups providing a more detailed insights into the grouping structure of firms in a network.  Thirdly, we see that hotel and pension is one of the major service sector in this area which could be confirmed by the fact that Hokkaido is one of the main sightseeing spot in Japan.  Finally, due to the fact that the presented approach only uses interfirm buyer--seller relations as the main network data, the presented approach fails to take into account the public and finance sector.  These two sectors are present in Figure 4a showing that the two approach shows complementary insights.

% IO
\begin{figure*}
	\begin{minipage}[b]{0.495\linewidth}
		\centering
		\includegraphics[width=1.03\linewidth]{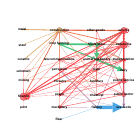}
		\subcaption{Network plot showing the traditional input--output table of the Nemuro-Kushiro area.  For ease of comaprison only strongly connected edges are depicted.  The size of the link reflects its strength.}
		\label{fig_A}
	\end{minipage}
	\begin{minipage}[b]{0.495\linewidth}
		\centering
		\includegraphics[width=1.17\linewidth]{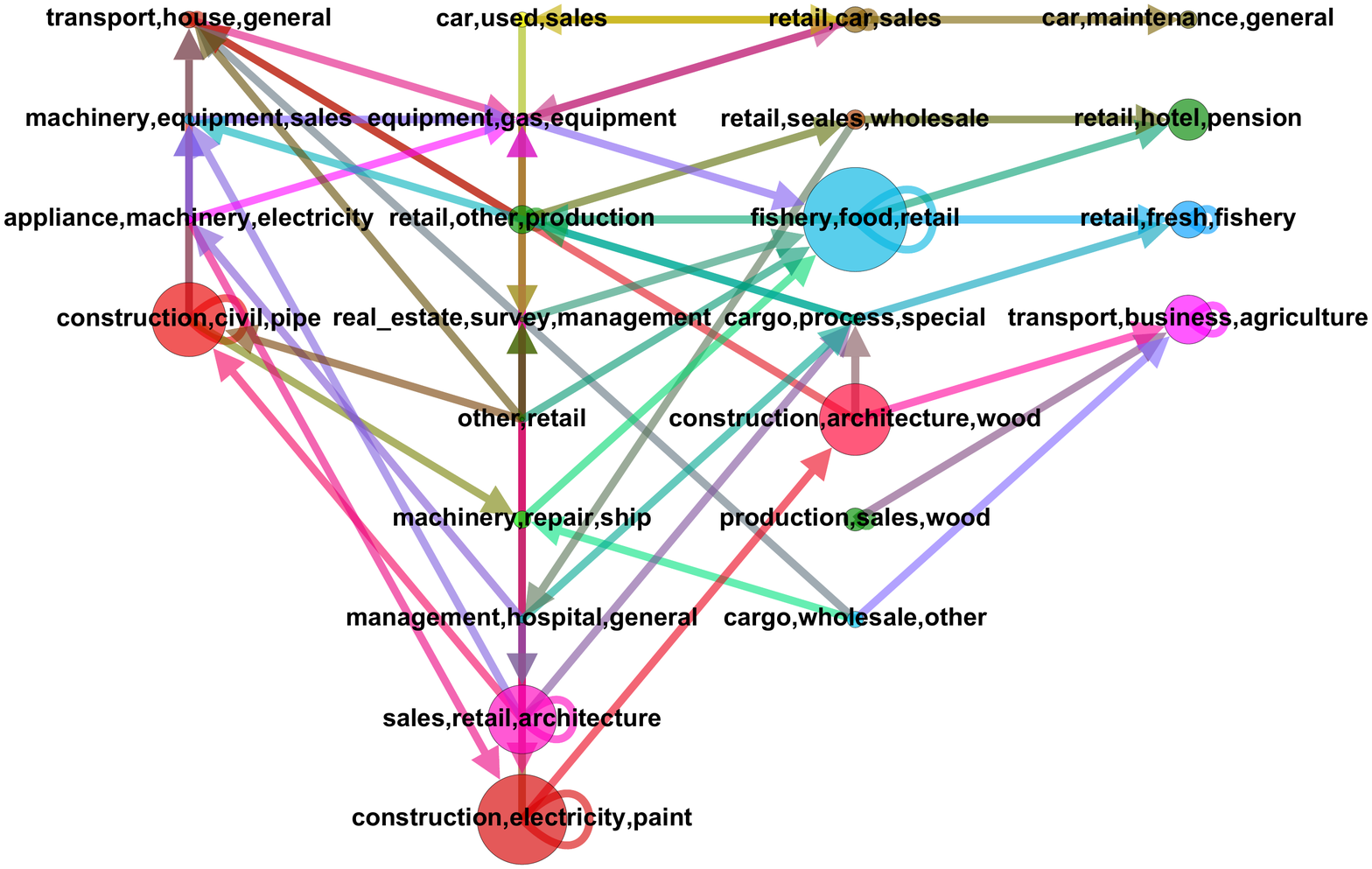}
		\subcaption{
			Network plot showing the the estimated interindustry structure of the Nemuro-Kushiro area using InfSBT.}
		\label{fig_B}
	\end{minipage}
	\caption{Comparison of our proposed approach to the traditional input--output table.  Node size is adjusted by their number of links.}
	\label{fig2D}
\end{figure*}

\section{Conclusions} 

% P1
Motivated by the practical problem of estimating the interdependence of industries in an economy, this paper first introduced InfSBT, a Bayesian nonparametric model of network formation that can (i) jointly model sparse network information and node textual information and (ii) jointly estimate the underlying latent block structure and number of components required to sufficiently represent the topology of a network.  The model is an extension of a previous SB model~\cite{Parkkinen2009}, which jointly models node textual information and an unbounded number of industries and interactions among them. The second aspect of the model was determined by defining a prior distribution that can process infinite mixtures in the network model.  For this task, we introduced the two-dimensional Chinese restaurant process, which builds on its famous one-dimensional counterpart.  We showed that, with sufficient approximation, the joint distribution derived from this process could be successfully used to define a model with an unbounded number of components.  We tested the model using both synthetic and real datasets.

% P2
By using the strength of dimensionality reduction, we determined the underlying interdependence of industries in a real-world network and outperformed previous models in predictive tasks.  There are other types of model that also use a dimensionality reduction approach to link formation \cite{Hoff2001}.  However, this type of latent space model cannot provide an interpretable summary of the underlying block structure and instead displays each node in an abstract space.  The proposed model provides a concise summary of the underlying block structure, as shown in Figure~\ref{fig4_2}.  Other works that jointly model the network link structure and node attributes focus on learning the node labels and are essentially not dimensionality reduction techniques~\cite{Moore2011,Bilgic2010}.  Hence, they cannot provide interpretable clusters and cannot respond to the underlying motivation of this paper.

%ACKNOWLEDGMENTS are optional
\section{Acknowledgments}
The author is grateful to Tsutomu Watanabe, Takayuki Mizuno, Takaaki Ohnishi, Yuichi Ikeda, Hiroshi Iyetomi, Shoji Fujimoto and RIETI for their helpful comments about the paper and for providing the data.  RH was supported by funding from the Research Fellowships of Japan Society for the Promotion of Science for Young Scientists. 

\bibliographystyle{abbrv}
\bibliography{Hisano_myref}  % sigproc.bib is the name of the Bibliography in this case

\end{document}